# FastONN – Python based open-source GPU implementation for Operational Neural Networks

Junaid Malik, Serkan Kiranyaz, *Member IEEE*, and Moncef Gabbouj, *Fellow, IEEE*

**Abstract.**

Operational Neural Networks (ONNs) have recently been proposed as a special class of artificial neural networks for grid structured data. They enable heterogenous non-linear operations to generalize the widely adopted convolution-based neuron model. This work introduces a fast GPU-enabled library for training operational neural networks, FastONN, which is based on a novel vectorized formulation of the operational neurons. Leveraging on automatic reverse-mode differentiation for backpropagation, FastONN enables increased flexibility with the incorporation of new operator sets and customized gradient flows. Additionally, bundled auxiliary modules offer interfaces for performance tracking and checkpointing across different data partitions and customized metrics.

## 1 Introduction

CNNs have advanced the state-of-the-art in challenging computer vision problems including object recognition, detection and segmentation [1]–[3]. Each of the stacked layers of convolutional neurons inside a CNN, applies filtering to local regions (receptive fields) of the input using the discrete convolution operation with learnable filter coefficients. Despite its wide-scale adoption, the convolutional neuron model, being an extension of a simple perceptron, is inherently linear, where the lone source of non-linearity is the point-wise non-linear activation. Consequently, a significantly large number of layers with interlaced non-linear activation functions are required to yield a hypothesis space strong enough to navigate complex non-linear spaces. Recently, the idea of Operational Neural Networks (ONN) [4] has been proposed which is a diversification of the convolutional neural model that involves embedding non-linear transformations at the level of individual receptive fields guided by learnable kernels.

The primary building block of the ONN framework is the *operational neuron model* which extends the principles of GOPs [5]–[7] to convolutional realm. While retaining favorable characteristics of sparse-connectivity and weight-sharing, an operational neuron provides the flexibility to incorporate non-linear transformations within local receptive fields without the overhead of additional trainable parameters. Driven by rich non-linear operators (operator sets), ONNs were shown to outperform CNN across a variety of challenging learning problems such as image-to-image translation, denoising, synthesis and segmentation [4]. However, the primary bottleneck hindering large scale adoption of ONNs is the lack of an open-source fast GPU enabled implementation which can facilitate reproducibility and fuel independent research efforts.

In this study, we propose FastONN, the first open-source Python-based and GPU-enabled ONN implementation, which aims at facilitating training and configuration of ONNs. FastONN leverages a novel formulation of an operational neuron which allows efficient utilization of GPU for increased computational efficiency. The library comprises of standalone modules, which allow training of end-to-end operational networks while preserving the ability to embed an operational module (operational layer/neuron) within any given network architecture. In addition, by automating gradient calculation, FastONN eases incorporation of novel operators to further diversify the operator set library.

## 2 Problems and Background

The primary building block of a convolutional neuron is the 2D discrete convolution operation. The convolution of a 2D image $y \in \mathbb{R}^{M \times N}$ with a filter $w \in \mathbb{R}^{m \times n}$ is given as:

$$x(i,j) = \sum_{u=0}^{m-1}\sum_{v=0}^{n-1} w(u,v)y(i-u, j-v) \qquad (1)$$

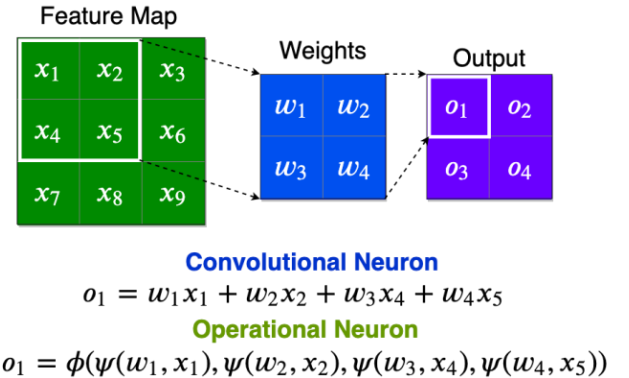

**Figure 1** Operational neuron incorporates patch-wise non-linear operations in the linear formulation of the convolutional neuron using generic nodal $\psi$ and pool $\phi$ functions.

Given any operator set $\theta = (\psi, \phi, f)$, an operational neuron equipped with $\theta$ modifies the operation of (1) as depicted in Figure 1 and formulated as:

$$x(i,j) = \phi_{u=0}^{m-1}\phi_{v=0}^{n-1}\left(\psi\big(w(u,v), y(i-u, j-v)\big)\right) \qquad (2)$$

where $\psi, \phi$ and $f$ are termed as nodal, pool and activation operator functions respectively.

FastONN uses an alternate formulation of the above operation. Firstly, $y$ is reshuffled such that values inside each $m \times n$ sliding block of $y$ are vectorized and concatenated as rows to form a matrix $Y \in \mathbb{R}^{\widehat{M} \times \widehat{N}}$ where $\widehat{M} = MN$ and $\widehat{N} = mn$. We can denote each element of $Y$ as,

$$Y(i,j) = y(\neg(i,j)) \quad (3)$$

where $\neg$ is a matrix that stores the index locations of $y$ in $Y$. This operation is commonly referred to as *"im2col"* and is critical in conventional GEMM-based convolution implementations. Secondly, we construct a matrix $W \in \mathbb{R}^{\widehat{M} \times \widehat{N}}$ whose rows are repeated copies of $\vec{w} = vec(W) \in \mathbb{R}^{mn}$, where $vec(\cdot)$ is the vectorization operator. Each element of $W$ is given by the following equation:

$$W(i,j) = \vec{w}(i) \quad (4)$$

The convolution operation of (1) can then be expressed as,

$$x = vec_{M \times N}^{-1}\left(\sum_j (Y \otimes W)\right) \quad (5)$$

where $\otimes$ represents the Hadamard product, $\sum_i$ is the summation across the dimension $i$. In (5), $vec_{M \times N}^{-1}$ is the inverse vectorization operation that reshapes the $MN$-sized vector back to $M \times N$. The alternate formulation of operational neuron in (3) can now be generically reformulated as follows:

$$x = vec_{M \times N}^{-1}(\phi(\psi(Y,W))) \quad (6)$$

where $\psi(\cdot): \mathbb{R}^{M \times N} \to \mathbb{R}^{M \times N}$ and $\phi(\cdot): \mathbb{R}^{\widehat{M} \times \widehat{N}} \to \mathbb{R}^{\widehat{M}}$ are the *nodal* and *pool* functions respectively. It can be clearly seen that the convolution operation of (5) is now a special form of (6) with nodal function $\psi(\alpha, \beta) = \alpha * \beta$ and pooling function $\phi(\cdot) = \sum_i$. To complete the forward-propagation, a non-linear activation operation $f$ follows the convolution. So, the final output of a convolutional neuron can be expressed as follows:

$$y = vec_{M \times N}^{-1} f(x) \quad (7)$$

where $f(\cdot)$ represents the point-wise activation operation. With the aforementioned formulation, each forward pass inside an operational neuron involves a single Hadamard product between two large matrices, $Y$ and $W$. Such a formulation is highly desirable for efficient utilization of GPU resources as it prevents intermittent memory copying and allows advanced optimization techniques such as tiling for enhanced computational performance [8]. Furthermore, as discussed in detail in Section 4.1, the backpropagation, and in turn the gradient calculation process, is handled by the automatic reverse-mode backpropagation engine, thus enabling differentiation through any choice of $\psi$, $\phi$ and $f$, as long as they follow the semantics provided in Table 2.

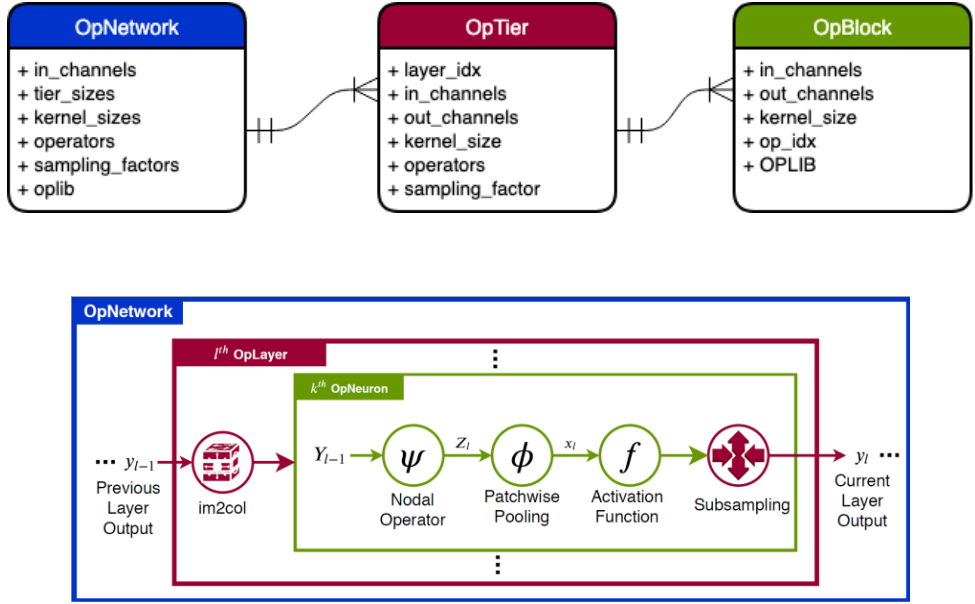

**Figure 2** Top: Entity Relationship Diagram (ERD) of the 3 core modules of FastONN. Bottom: Aggregation of FastONN modules to configure an ONN model. The formulation shown is for the $k^{th}$ neuron of the $l^{th}$ layer of the network.

## 3 Software Framework

### 3.1 Software Architecture

The core of the library consists of three main classes, as depicted in the entity-relationship diagram in Figure 2 and described below:

o **Op**erational **Block**: class implementing the atomic operational neuron. Each OpBlock object has an associated operator set that is used for accomplishing the required non-linear operation.
o **Op**erational **Tier**: has one or more operational blocks and implements the unfolding operation, propagation through operational blocks and subsampling operations. An OpTier object represents the abstraction of a neural network layer and can be used as an out-of-the-box replacement for a convolutional layer in any given architecture.
o **Op**erational **Network**: has one or more operational tiers and implements propagation through the layers. An OpNetwork object is an end-to-end ONN that can be trained in the same way as any other ANN model.

In addition to the main classes listed above, a variety of auxiliary modules are also provided that facilitate training:

o OSL: operator set library module which houses all nodal, pool and activation function definitions. Table 2 shows details of a sample operator set library along with generic formulations for nodal, pool and activation functions that can be used to add custom functions.
o Trainer: helper class which supports a wide array of functionalities used for training a model for the given number of epochs and randomly initialized runs, evaluation on validation and test data, plotting of metrics, performance tracking and model checkpointing across all metrics and data partitions.

### 3.2 Software Functionalities

Following are the major highlights of the features of FastONN:

- Configuration of a fully operational end-to-end network with customized number of layers, neurons per layer, kernel sizes and subsampling factors. Each layer is capable of full heterogeneity with the ability of specifying a different operator set for individual neurons.
- A comprehensive operator set library with multiple predefined nodal, pool and activation functions. New operators can be added following simple semantics as illustrated in Table 2.
- Back-Propagation (BP) training an ONN network on GPU using custom defined objective functions and any choice of major optimization techniques such as SGD with momentum, Adam etc.
- Defining custom performance metrics and dynamically tracking them across the training process for all data partitions. Automatic checkpointing is enabled for each metric to allow rolling back of the model to any desired state. Any state of the model is not limited to the values of trainable parameters, but also includes the state of the optimizer and other training statistics in order to allow seamless resumption of the training.

## 4 Implementation and Empirical Results

### 4.1 Implementation details

FastONN is based on PyTorch [9], a popular Python-based deep learning framework, which enables imperative Pythonic programming styles for designing high-performance GPU-accelerated neural networks. In addition, Pytorch leverages the AutoGrad [10] package to provide an intuitive interface for reverse-mode automatic differentiation, thus automating the gradient computation process [11] for many choices of nodal, pool and activation functions. Nevertheless, the ability to define custom backpropagation formulation is still preserved and can be used to define custom derivatives and perform gradient modification for complex operators. This is essential for a flexible ONN implementation as the gradient flows within operational neurons are highly customized [4], owing to patch-wise non-linearities. A detailed comparison of analytical and automatic backpropagation through an operational neuron is provided in the supplementary material. PyTorch also supports broadcasting semantics which allow making virtual copies of a tensor along certain dimension(s) without explicitly copying the data, which results in improved memory utilization. This is essential for an operational neuron as it provides a memory-efficient way to accomplish the formulation of (3). All the standalone modules of FastONN shown in Figure 2 inherit from the `nn.Module` base class of Pytorch, thus ensuring compatibility and reusability. The source-code for FastONN is available online as a public Github repository and uses Git versioning system as well as supporting continuous integration with automated build and tests.

### 4.2 Empirical Results

We provide empirical results of training an ONN for the image transformation as introduced in [4], using the FastONN library. The task involves configuring and training an ONN for learning 10 distinct folds of 4-to-4 face image translations. Table 1 presents the results for each of the 10 folds along with the computational efficiency in terms of time taken per image to train the network on GPU, including performance tracking, plotting and checkpointing.

## 5 Illustrative Examples

In this section, we provide the code illustrations for the experiments conducted to produce the results presented in

**Table 1. Fold-wise performance w.r.t SNR (dB) and per image training time on GPU for Image Transformation problem using FastONN library.**

| Fold 1 | Fold 2 | Fold 3 | Fold 4 | Fold 5 | Fold 6 | Fold 7 | Fold 8 | Fold 9 | Fold 10 | Mean | GPU Time (s) |
|---|---|---|---|---|---|---|---|---|---|---|---|
| 16.5 | 16.4 | 9.26 | 16.4 | 12.2 | 12.1 | 16.7 | 10.3 | 15.6 | 14.4 | 13.98 | 0.4375 |

**Table 2. A subset of the operator set library providing mathematical expressions for each operator and its corresponding implementation in code.**

| | | **NODAL FUNCTIONS ($\psi$)** | |
|---|---|---|---|
| *Generic* | $\psi(w, x)$ | Any element-wise operation between x[:,:,None,:,:] and w[None,:,:,:,None] | |
| Multiplication | $wx$ | `def mul(x,w): return x[:,:,None,:,:].mul(w[None,:,:,:,None])` | |
| Sinusoid | $\sin(K_s w x)$ | `def sine(x,w,K_SIN): return torch.sin(K_SIN*mul(x,w))` | |
| Exponential | $e^{wx}$ | `def expp(x,w): return (torch.exp(mul(x,w)) - 1)` | |
| Chirp | $\sin(K_c w x^2)$ | `def chirp(x,w,K_CHIRP): return (torch.sin(K_CHIRP*mul(x.pow(2),w)))` | |
| | | **POOL FUNCTIONS ($\phi$)** | |
| *Generic* | $\phi(Z_l)$ | Any reduction function along 3$^{rd}$ dimension | |
| Summation | $\sum Z_l$ | `def summ(x): return torch.sum(x,dim=3)` | |
| Median | $median(Z_l)$ | `def medd(x): return (x.shape[2]*torch.median(x,dim=3))[0]` | |
| Max | $\max(Z_l)$ | `def maxx(x): return (x.shape[2]*torch.max(x,dim=3))[0]` | |
| | | **ACTIVATION FUNCTIONS ($f$)** | |
| *Generic* | $f(x)$ | Any pointwise non-linear function b/w x and b[None,:,None,None] | |
| Tangent Hyperbolic | $\tanh(x_l) = \dfrac{1 - e^{-2x}}{1 + e^{-2x}}$ | `def tanh(x,b): return torch.tanh(x-b[None,:,None,None])` | |
| Linear Cut | $lincut(x, cut) = \begin{cases} \dfrac{x}{cut} & \text{if } abs(x) \leq cut \\ sign(x) & \text{if } abs(x) > cut \end{cases}$ | `def lincut(x,b,cut=10): return torch.clamp((x-b[None,:,None,None])/cut, -1,1)` | |

Section 4.2. In the code snippet of Figure 3, the **getOPLIB** utility function is used to generate an operator set library using the desired nodal, pool and activation functions. The mathematical formulation and the corresponding code implementation for each of the operator set used is provided in Table 2. This is followed by instantiating an **OpNetwork** object with the required network architecture parameters. For each of the operational layers in the network, operator set indices are passed to the **operators** attribute of the **OpNetwork** object. The **model** variable now holds a fully configured ONN with the desired number and type of operational layers and neurons and can be used as any other Pytorch model.

In Figure 4, a sample code is provided illustrating how the Trainer class, which comes bundled with FastONN, can be used to train the ONN model with any choice of objective functions, optimization technique, learning rate, performance metrics and device. The **metrics** attribute takes a Python dictionary as input with metric name and calculation rule as key value pairs. Each metric's calculation rule is a Python tuple which contains the function used to calculate the metric's value, along with the criteria used for tracking the best state of the metric. In the example of Figure 4, the Trainer object would track the SNR value ('snr') and save the state of the model which maximizes the SNR for each data partition. Tracking of the loss function is performed by default. Furthermore, the **save_all** method is used to save all the statistics and model checkpoints in a single .pth file.

```
# Definition of operator set library
OPLIB = getOPLIB(
    NODAL = [mul,cubic,sine,expp,sinh,chirp],
    POOL = [tanh,lincut],
    ACTIVATION = [summ,lincut]
)
# Configuration of a heterogenous ONN
model = OpNetwork(
    in_channels=1, # number of input channels
    tier_sizes=[12,32,1], # number of neurons in hidden and output layer
    kernel_sizes=[21,7,3], # kernel size
    operators=[ # operators for each layer
        [1], # Hidden Layer 1
        [6], # Hidden Layer 2
        [3] # Output Layer
    ],
    sampling_factors = [2,-2,1], # scaling factors for each layer
    OPLIB=OPLIB, # operator set library
)
```

**Figure 3** Code snippet illustrating the configuration of an ONN model using FastONN library.

```
trainer = Trainer(
    model=model, # ONN model to train
    train_dl=train_dl, # training dataloader
    val_dl=val_dl, # validation dataloader
    test_dl=test_dl, # test dataloader
    loss=torch.nn.MSELoss(), # loss function
    opt_name='cgd', # optimization technique
    lr=0.001, # initial learning rate
    metrics={'snr':(calc_snr,'max')}, # metrics to calculate and track
    device='cuda', # device used for training
    reset_fn=model.reset_parameters, # method used to reset network parameters
    model_name="image_transformation" # model name
)

trainer.train(num_epochs=240,num_runs=3) # train model
trainer.plot_stats() # plot stats
trainer.save_all() # save all
```

**Figure 4 Code snapshot showing how the trainer utility class is used to train an ONN model.**

# 6   Conclusions

In this paper, we introduce FastONN, the first open-source and GPU enabled Python-based implementation of Operational Neural Networks (ONNs). Based on a novel reformulation of the operational neuron, FastONN provides an easy-to-use interface to configure and train ONNs as end-to-end models, embed operational layers in contemporary network architectures and integrate novel application specific non-linear operator sets. This paves the way for independent research efforts and makes it easy to implement off-the-shelf ONN solutions for industrial and academic use. Future software updates will comprise of additional modules for optimal operator search and hyperparameter optimization.

# 7   B- Required Metadata

## 7.1   B1 Current executable software version

**Table 3 Software metadata**

| Nr | (executable) Software metadata description | |
|---|---|---|
| S1 | Current software version | 0.1.1 |
| S2 | Permanent link to executables of this version | https://github.com/junaidmalik09/fastonn/releases/tag/0.1.1 |
| S3 | Legal Software License | MIT |
| S4 | Computing platform / Operating System | Linux, OSX, Unix-like, Windows. |
| S5 | Installation requirements & dependencies | Python, Anaconda, PyTorch |
| S6 | If available Link to user manual - if formally published include a reference to the publication in the reference list | https://github.com/junaidmalik09/onn_torched/README.md |
| S6 | Support email for questions | junaid.malik@tuni.fi |

## 7.2   B2 Current code version

**Table 4 Code metadata**

| Nr | Code metadata description | *Please fill in this column* |
|---|---|---|
| C1 | Current Code version | 0.1.1 |
| C2 | Permanent link to code / repository used of this code version | https://github.com/junaidmalik09/fastonn/releases/tag/0.1.1 |
| C3 | Legal Code License | MIT |
| C4 | Code Versioning system used | Git |
| C5 | Software Code Language used | *Python* |
| C6 | Compilation requirements, Operating environments & dependencies | *torch, torchvision, numpy, scipy, PIL, glob, matplotlib, pathlib* |
| C7 | If available Link to developer documentation / manual | *https://github.com/junaidmalik09/fastonn/README.md* |
| C8 | Support email for questions | *junaid.malik@tuni.fi* |